\documentclass[letterpaper, 10 pt, journal]{IEEEconf}  

\IEEEoverridecommandlockouts                              

\usepackage{graphics} 
\usepackage{epsfig} 
\usepackage{times} 
\usepackage{amsmath} 
\usepackage{amssymb}  
\usepackage{amsfonts}
\usepackage{subfigure}
\usepackage{algorithm}
\usepackage{algorithmic}
\usepackage{cite}

\title{\LARGE \bf
Learning from Local Experience: Informed Sampling Distributions \\ for High Dimensional Motion Planning
}

\author{Keita Kobashi$^{1}$, Changhao Wang$^{1}$, Yu Zhao$^{2}$, Hsien-Chung Lin$^{2}$, and Masayoshi Tomizuka$^{1}$
\thanks{*This work was partially supported by KDDI Foundation}
\thanks{$^{1}$Keita Kobashi, Changhao Wang, and Masayoshi Tomizuka are in Mechanical Engineering,
        University of California, Berkeley, CA, 94704
        {\tt\small \{kkobashi, changhaowang, tomizuka\}@berkeley.edu}}%
\thanks{$^{2}$Yu Zhao and Hsien-Chung Lin are with FANUC Advanced Research Lab,
        Kohoutec Way, CA 94587, USA
        {\tt\small \{Yu.Zhao, Hsien-chung.Lin\}@fanucamerica.com}}%
}

\begin{document}

\maketitle
\thispagestyle{empty}
\pagestyle{empty}

\begin{abstract}
This paper presents a sampling-based motion planning framework that leverages the geometry of obstacles in a workspace as well as prior experiences from motion planning problems. 
Previous studies have demonstrated the benefits of utilizing prior solutions to motion planning problems for improving planning efficiency.
However, particularly for high-dimensional systems, achieving high performance across randomized environments remains a technical challenge for experience-based approaches due to the substantial variance between each query.
To address this challenge, we propose a novel approach that involves decoupling the problem into subproblems through algorithmic workspace decomposition and graph search. 
Additionally, we capitalize on prior experience within each subproblem. 
This approach effectively reduces the variance across different problems, leading to improved performance for experience-based planners.
To validate the effectiveness of our framework, we conduct experiments using 2D and 6D robotic systems. 
The experimental results demonstrate that our framework outperforms existing algorithms in terms of planning time and cost. 


\end{abstract}

\section{INTRODUCTION}
Sampling-based motion planning is effective for many real-world applications such as navigation and robotic manipulation.
Traditionally, the sampler does not consider any environmental information that draws samples uniformly in configuration space. 
As a result, the number of valid samples is limited due to obstacles and systems constraints, and the planning process is inefficient.
To tackle this problem, many prior works adopt non-uniform sampling to prioritize drawing valid samples \cite{boor1999gaussian, hsu2003bridge, wilmarth1999maprm, rodriguez2006obstacle}.
Especially, the experience-based approach, which utilizes prior solutions to motion planning problems to estimate the feasible path of new motion planning problems achieves remarkable success \cite{ichter2018learning, kumar2019lego, chamzas2019using, chamzas2021learning, chamzas2022learning}.
However, as discussed in previous works, the accuracy of estimation can be low when learning randomized queries due to the high variance between each problem \cite{kumar2019lego, chamzas2019using, chamzas2021learning, chamzas2022learning}.
To tackle this technical challenge, we then consider decoupling a motion planning problem into simple subproblems to reduce the variance between each problem and using a non-uniform sampling distribution estimated by a generative model.
In this work, we adopt workspace decomposition as a means of decoupling the problem because decomposition process of a configuration space is generally heuristic and the time demanding grows up exponentially as the dimension increases even for relatively simple environments \cite{deits2015computing, marcucci2022motion}.
On the other hand, the dimension of workspace is at most 3, hence, the time demanding for decomposition process does not increase even if the robot has high degree of freedom (DoF).

\begin{figure*}[t]
 \centering
   \includegraphics[clip, width=2.0\columnwidth]{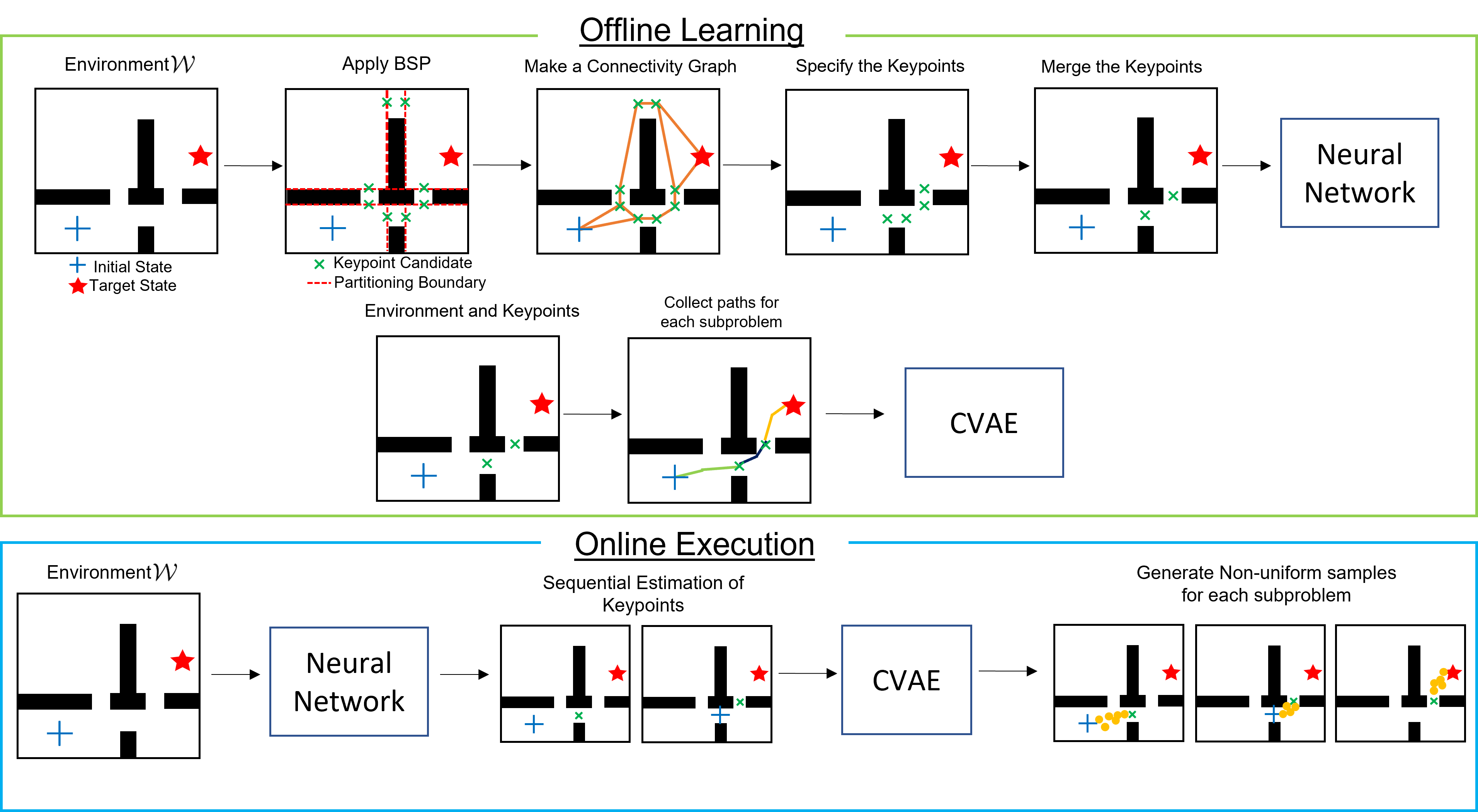}
 \caption{The overview of our motion planning framework. For a given environment (workspace), we decompose it by BSP and make convex subspaces that do not comprise obstacles. We assign keypoint candidates on the collision-free parts of partitioning boundaries represented as red dash lines in the figure. We then connect the keypoint candidates with straight lines since each subspace is convex and make a connectivity graph. Assigning weight (length of an edge) to each edge of the graph, we can specify the keypoint by simply employing classical graph search algorithm such as Dijkstra or A*. We merge the keypoints if the distance between each keypoint is less than a threshold. We pass these keypoint information to a neural network to train. We also train CVAE by using solutions to subproblems that are defined from a keypoint to an adjacent keypoint. For online execution, the trained neural network estimates keypoints sequentially for a given environment and CVAE generates non-uniform samples based on the estimated keypoints. We discuss the details of this framework in Section IV.}
 \label{framework_overview}
\end{figure*}
There are some previous works that address unifying workspace decomposition and experience-based non-uniform sampling \cite{chamzas2019using, chamzas2021learning, chamzas2022learning}.
The authors of these works are pioneers in the field and their approach succeeds in reducing the planning time compared to other experience-based planners.
They retrieve prior solutions from database based on the similarity of environments and make a distribution using Gaussian Mixture Models (GMM) using the solutions.     
However, they handle similar shelf environments and they do not present any general workspace decomposition scheme.
Besides, they do not propose the general way to evaluate the similarity from geometry of obstacles in a workspace.
Therefore, problems related to how to decompose a workspace and how to evaluate the similarity still remain.

Figure \ref{framework_overview} displays the overview of our planning framework.
In contrast to the prior works, as an algorithmic decomposition scheme, we employ Binary Space Partitioning (BSP) \cite{fuchs1980visible}.
BSP is one of the space partitioning algorithms that recursively subdivides a Euclidean space into two convex subspaces using hyperplanes as boundaries.
Compared to the prevailing cell decomposition, BSP can effectively use geometry of obstacles in a workspace by choosing the partitions from facets of obstacles.
This allows us to extract regions surrounded by obstacles.
In such regions, the number of valid samples (configurations) is limited and we call them challenging regions.
We can decouple a problem into subproblems using keypoints assigned in the challenging regions.
We specify the keypoints to use for decoupling a problem by graph search.
The authors of \cite{marcucci2022motion} integrate graph search into optimization-based motion planning framework. 
Then, using the solutions to subproblems, a sampler can draw the adequate number of samples in the challenging regions.
In this work, we employ Conditional Variational Autoencoder (CVAE) to utilize prior solutions.
CVAE can take environmental information as conditions and generate non-uniform samples around estimated optimal paths by considering the conditions \cite{ichter2018learning}, hence, we do not need to evaluate the similarity.
We believe other generative models can also be applicable to our framework.
To reduce the execution time of BSP, we employ a neural network to learn the keypoints.
Thus, the neural network directly estimates the keypoints.
Then we collect solutions of each subproblem to train CVAE using estimated keypoints.
We discuss the details of this in Section IV.

Finally, we conduct some experiments to demonstrate the effectiveness of our framework.
We execute our motion planning framework in 2D maze and 3D shelf picking environments with a free-flying robot and a 6-DoF manipulator.
The results of experiments indicate that our framework outperforms other state of the art planners in terms of planning time and cost.
Note that in this study, cost means path length.

The contributions of this paper are as follows:
\begin{itemize}
    \item We propose a novel motion planning framework that unifies Binary Space Partitioning and experience-based non-uniform sampling.
    \item We improve the accuracy of estimation from CVAE by reducing the variance between each motion planning problem.
    \item Numerical experiments demonstrate that our framework is robust to randomization and outperforms other state of the art planners in terms of planning time and cost. 
\end{itemize}

\section{RELATED WORK}
This work leverages the advantages of non-uniform sampling and local sampling through workspace decomposition.
In the following subsections, we mention earlier work related to the above-mentioned research field.

\subsection{Local Sampling through Workspace Decomposition}
Enormous studies have discussed workspace decomposition to accelerate the planning process.
Traditionally, cell decomposition \cite{plaku2010motion, scheurer2011path, abbadi2015safe} and triangular mesh \cite{bhatia2010motion} are used to decompose a workspace.
Besides, the authors of \cite{brock2001decomposition} use balls to decompose a workspace.
However, in these work, they decompose a workspace without using the geometry of obstacles effectively.
Hence, it is difficult to specify challenging regions from the decomposition results algorithmically.
On the other hand, BSP takes geometry  of obstacles in a workspace into account in the decomposition results, thus we can easily detect the challenging regions in the workspace from the decomposition results.
In this work, we use BSP to decompose a workspace and specify the key points.
BSP is used in the motion planning context first in \cite{tokuta1991motion} for 2D motion planning problems.
Then, the authors of \cite{mesesan2018hierarchical} apply BSP to planer manipulator scenarios.
However, these works apply BSP only to simple environments and do not consider any complex environments that will make challenging regions in a workspace.
In addition, the calculation cost of BSP is $O(|\mathcal{F}|^2)$ where $\mathcal{F}$ is a set containing all facets of obstacles in a workspace and $|\mathcal{F}|$ is its cardinality \cite{paterson1990efficient}.
Therefore, if the workspace is complex, BSP execution may be a bottleneck of the entire planning process.
In our work, we utilize neural network structure to learn the partitioning by BSP and reduce the execution time for BSP.

\subsection{Non-Uniform Sampling}
Most fundamental approach to draw samples from a configuration space is uniform sampling that draws samples with equal probability.
Although uniform sampling has probabilistic completeness i.e. all possible samples are drawn after infinite iterations, the number of valid samples for a given motion planning problem is usually limited.
Hence, only using uniform sampling leads to inefficient sampling process and previous research tried to reasonably bias the sampling distribution to realize non-uniform sampling.

Prior work has realized non-uniform sampling using workspace information \cite{boor1999gaussian, hsu2003bridge, wilmarth1999maprm, rodriguez2006obstacle}.
One way is generating dense samples in challenging regions by regular decomposition in a workspace.
\cite{van2005using} uses octree to extract challenging regions and generates dense samples in the narrow regions to guide samplers.
\cite{kurniawati2004workspace, plaku2015region} use triangular meshes to extract challenging regions and realize dense sampling.  
\cite{kallman2004motion} utilizes workspace decomposition to create a dynamic roadmap for changing environments.
These approaches are effective for low-dimensional systems in relatively simple environments because it is easier to specify the corresponding configurations for these systems from workspace information.
However, for high-dimensional systems, such as robotic manipulators, snake robots, and legged robots, it can be difficult and time consuming to specify the configuration online due to the inverse kinematics problems.
Besides, these works do not transfer the experience to other similar problems.
To handle this issue, our framework utilizes a neural network structure and CVAE to directly estimate corresponding configurations from workspace information.

In recent years, learning technique achieved remarkable success to accelerate the planning process.
\cite{wang2020neural, chen2019learning, qureshi2019motion, terasawa20203d, jurgenson2019harnessing, yonetani2021path, bency2019neural} utilize neural network to guide planners.
Another approach is generating non-uniform samples in a configuration space by learning from prior solutions of motion planning problems \cite{ichter2018learning, ichter2020learned, kumar2019lego, zhang2018learning, gaebert2022learning, chamzas2019using, chamzas2021learning, chamzas2022learning}.
The drawback of these approaches is difficult to apply for general cases because, particularly for high dimensional systems, the variance between each query can be large even if the difference in a workspace is small as discussed in \cite{chamzas2021learning}.
\cite{chamzas2019using, chamzas2021learning, chamzas2022learning} partially overcome this drawback using database approach.
However, the performance of the planner depends on the data collected before planning.
If the positions of obstacles are changed, the prior solutions do not work and if we make the planner robust to randomization, we have to store plenty of solutions. 
Thus this approach is still vulnerable to workspace difference.
Apart from this remaining issue, the execution time for workspace decomposition can be a problem if the environment becomes complex.

In our framework, we decouple a motion planning problem into simple subproblems using BSP and this leads to reduction of variance among different problems.
This improves the estimation quality of CVAE and thus we can obtain non-uniform samples accurately in challenging regions.
Besides, we learn the workspace decomposition using a neural network, hence, our approach is effective even in complex environments.

\section{PROBLEM FORMULATION}
In this section, we discuss motion planning problems we handle in this study and the details of BSP and CVAE.

\subsection{Problem Statement}
The objective of a motion planning problem is to find a collision-free path.
Let $\mathcal{C} \subset \mathbb{R}^n$ be a configuration space of a robot, and let $\mathcal{C}_\mathrm{col}$ be a collision region. 
$\mathcal{C}_\mathrm{free} = \mathcal{C} \backslash \mathcal{C}_\mathrm{col}$ denotes a collision-free region in the configuration space.
Initial configuration $x_{\mathrm{init}}$ and target configurations $\mathcal{C}_{\mathrm{target}}$ must be in $\mathcal{C}_{\mathrm{free}}$.
Let $\xi (t)$ denote a path and this is a continuous function that maps a parameter $t \in [0,1]$ to a configuration in $\mathcal{C}$.
Using these notations, the objective can be described as finding a continuous map $\xi(t)$ that satisfies $\xi(t) \in \mathcal{C}_\mathrm{free}, ^\forall t \in [0,1]$, $\xi(0) = x_{\mathrm{init}}$, and $\xi(1) \in \mathcal{C}_{\mathrm{target}}$.

In this work, we mainly use Rapidly-exploring Random Tree (RRT) \cite{lavalle2001randomized} and its variants \cite{karaman2011sampling}, \cite{kuffner2000rrt} as motion planners.

\subsection{Binary Space Partitioning}
As mentioned in the previous section, BSP is one of the partitioning algorithms that divides a real Euclidean space into convex subspaces.
\begin{figure}[t]
 \centering
    \subfigure[Triangle]{%
        \includegraphics[clip, width=0.4\columnwidth]{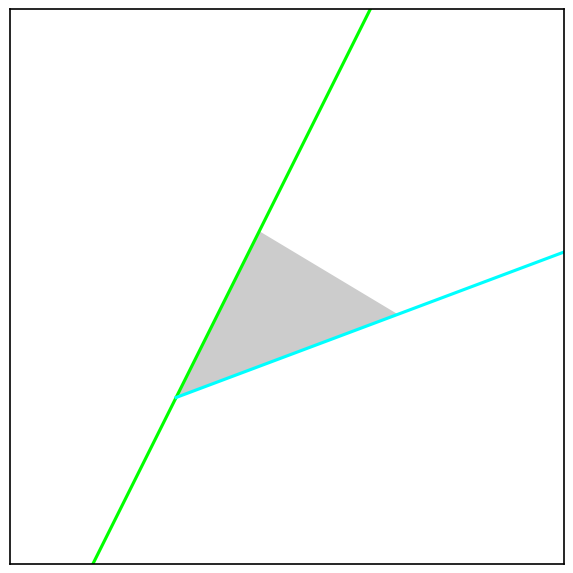}}
    \subfigure[Maze]{%
        \includegraphics[clip, width=0.4\columnwidth]{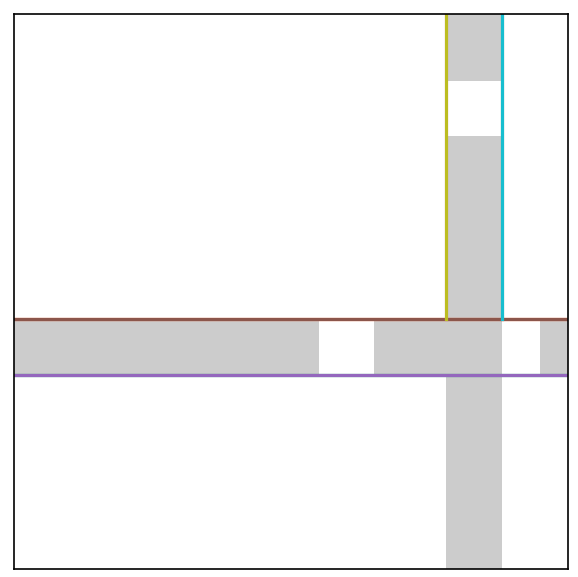}}
 \caption{Examples of BSP. The environments are subdivided into convex subspaces. We can see the boundaries do not intersect with each other and automatically extract challenging regions.}
 \label{BSP}
\end{figure}
Figure \ref{BSP} exhibits examples of partitioning.
In the motion planning context, we choose partitioning boundaries from facets of obstacles.
This enables us to extract the challenging regions by picking up the collision-free parts of the boundaries. 
The advantages of BSP in motion planning are 1) each subspace becomes convex and 2) automatically extracts challenging regions in a workspace.
We discuss how we utilize these advantages in Section IV.
Thus, BSP ensures every subspace does not comprise any obstacles and any segments connecting two points in each subspace are collision-free.

\subsection{Conditional Variational Autoencoder}
In this work, we utilize CVAE to generate a probability distribution whose extremums are around estimated collision-free paths in subspaces.

Let $\mathit{x}$ denote a configuration of a robot and $\mathcal{W}$ denote a condition, in this case it is equivalent to initial and target states and position of obstacles.
The objective of CVAE training is estimating the probability distribution $p_{\theta^*}(x|\mathcal{W})$ that describes the probability of the states around optimal paths where $\theta^*$ denotes a set of true parameters.
The probability $p_\theta (x|\mathcal{W})$ can be written in the following equation.
\begin{align}
    \int p_\theta (x|z,\mathcal{W}) p_\theta (z|\mathcal{W}) dz
\end{align}
Besides, we determine $\theta^*$ by maximum likelihood estimation, thus
\begin{align}
    \theta^* = \arg \max_\theta \frac{1}{N} \sum_{i=1}^N \log \left(\int p_\theta (x_i|z,\mathcal{W}) p_\theta (z|\mathcal{W}) dz \right)
\end{align}
However, the calculation of the integral part is completely intractable.
Therefore, we introduce a tractable function $q_\phi (z|x,\mathcal{W})$ as an encoder for approximation and maximize the Evidence Lower Bound instead
\begin{equation}
    \begin{aligned}
        \log ( p_\theta (x|\mathcal{W}) ) \geqq & -D_{KL} (q_\phi (z|x,\mathcal{W}) \| p(z|\mathcal{W})) \\ & +  
        \mathbb{E}_{z \sim q_\phi (z|x,\mathcal{W})} [\log (p_\theta (x|z,\mathcal{W}))]
    \end{aligned}
\end{equation}
where $D_{KL}$ denotes Kullback-Leibler divergence.
In this work, we utilize encoder-decoder network of CVAE using neural network. 
Inside the CVAE structure, encoder draws a latent variable $\mathit{z}$ with probability of $q_\phi (z|x,\mathcal{W})$.
Decoder draws sample $\mathit{x}$ with the probability of $p_\theta (x|z,\mathcal{W})$.
The prior of the probability $p_\theta (z|\mathcal{W})$ is generally fixed to a multivariate standard Gaussian distribution $\mathcal{N} (0|I)$.

\section{LEARNING FROM LOCAL EXPERIENCE}
This section presents our novel motion planning framework.
Our motion planner possesses two neural network structure.
The first neural network sequentially estimates keypoints in challenging regions in a workspace.
Using these keypoints, we can divide a motion planning problem into subproblems by treating each keypoint as subgoal.
The second neural network is CVAE and CVAE biases the probability distribution to prioritize the sampling process in subproblems that are defined between each keypoint. 

The following subsections discuss the details of its training and execution procedures in this section.

\subsection{Offline Learning (Algorithm \ref{alg_offline_learning})}
Algorithm \ref{alg_offline_learning} describes the offline learning procedure of our motion planner.
To train the neural network and CVAE, we collect training data related to a sequence of keypoints using BSP, and paths using RRT or its variants (line 1).
We then train the neural network using keypoints data (line 2).
As the first step to train, we pick up the first keypoint from the sequence and pass it as label. 
Then, we pass the environment information, initial state, and target state as data so that the neural network becomes able to predict the next keypoint.
After this process, we pick up the second keypoint from the sequence and treat the first keypoint as an initial state to train the neural network.
Through this training procedure, the neural network can predict keypoints sequentially.
For the better demonstration, we enforce the initial and target states in the data as the size of them becomes equal to the size of environment information. 
We also train CVAE by feasible solutions of subproblems (line 3).
As recommended in \cite{ichter2018learning}, we train the CVAE conditioned on initial and target states in addition to environment information.
Hence, in this case, we train the CVAE sequentially by treating keypoints as subgoals.

\begin{algorithm}[t]
\small 
\algsetup{linenosize=\small}
\caption{Offline Learning}
\begin{algorithmic}[1] 
\label{alg_offline_learning}
\REQUIRE initial state $x_{\mathrm{init}}$, target state $x_{\mathrm{target}}$, environment $\mathcal{W}$
\STATE $k^*_1, \ldots, k^*_l, \mathrm{paths} \gets \mathrm{DataCollection} (x_{\mathrm{init}}, x_{\mathrm{target}}, \mathcal{W})$ 
\STATE TrainNeuralNetwork($k^*_1, \ldots, k^*_l, x_{\mathrm{init}}, x_{\mathrm{target}}, \mathcal{W}$)
\STATE TrainCVAE(paths, $k^*_1, \ldots, k^*_l, x_{\mathrm{init}}, x_{\mathrm{target}}, \mathcal{W}$)
\end{algorithmic}
\end{algorithm}
\begin{algorithm}[t]
\small 
\algsetup{linenosize=\small}
\caption{Training Data Collection}
\begin{algorithmic}[1] 
\label{alg_data_collection}
\REQUIRE initial state $x_{\mathrm{init}}$, target state $x_{\mathrm{target}}$, environment $\mathcal{W}$
\STATE $\mathcal{L}_1, \ldots, \mathcal{L}_N \gets \mathrm{BSP} (\mathcal{W})$ 
\STATE $k_1, \ldots, k_m \gets \mathrm{GetKeyPointCandidates} (\mathcal{L}_1, \ldots, \mathcal{L}_N)$
\STATE $\mathcal{G} \gets \mathrm{GetConnectivityGraph} (\mathcal{L}_i's, k_i's, x_{\mathrm{init}}, x_{\mathrm{target}}$) 
\WHILE{isFeasible == False}
    \STATE $k^*_1, \ldots, k^*_l \gets \mathrm{RunDijkstra}(\mathcal{G})$
    \STATE $\mathrm{isFeasible}, \ \mathrm{path} \gets \mathrm{RunRRT} (k^*_1, \ldots, k^*_l)$
    \IF{isFeasible == False}
    \STATE $\mathrm{RemoveEdge}(\mathcal{G}, k^*_i)$ 
    \ELSE
    \STATE $\mathrm{isPathCollisionFree} \gets \mathrm{ExaminePath}(\mathrm{path})$
    \IF{$\mathrm{isPathCollsionFree} == \mathrm{False}$}
    \STATE $\mathrm{RemoveEdge}(\mathcal{G}, k^*_i)$ 
    \STATE isFeasible $\gets \mathrm{False}$
    \ENDIF
    \ENDIF
\ENDWHILE
\FORALL{$k^*_i \in \{k^*_1, \ldots, k^*_l\}$}
    \IF{$\| k^*_{i+1} - k^*_i \|_2 < \varepsilon$}
    \STATE $k^*_{i+1} \gets ( k^*_{i+1} + k^*_i) /2$ 
    \STATE Remove($k^*_i$)
    \ENDIF
\ENDFOR
\STATE keypoints $\gets k^*_1, \ldots, k^*_l$ 
\RETURN keypoints, path
\end{algorithmic}
\end{algorithm}
\subsection{Training Data Collection (Algorithm \ref{alg_data_collection})}
Algorithm \ref{alg_data_collection} explains the procedure of data collection to train the neural network and CVAE.
In the data collection process, we firstly decompose a workspace and obtain keypoint candidates (lines 1--3).
Then, we specify the keypoints from the candidates by using a graph (lines 4--22).
We discuss the details in the following part of this section.

To begin with, we execute BSP using the workspace information to decompose a workspace and obtain convex subspaces $\mathcal{L}_1, \ldots, \mathcal{L}_N$ (line 1).
Then, we assign key point candidates $k_1, \ldots, k_m$ on the boundaries created by BSP(line 2).
The boundaries are hyperplanes of a given workspace, such that lines in a 2D environment or planes in a 3D environment.
Using this setting, we become able to generate dense non-uniform samples by CVAE that utilizes two adjacent keypoints as substart and subgoal.
Although we can freely assign the resolution of pixel grid, we recommend the number of pixels in a row becomes odd number because the original middle point is included.
In line 3, we create a connectivity graph of each keypoint.
Since we choose partitioning boundaries from facets of obstacles and BSP creates convex regions, a line segment of any two points in a subspace must be in the subspace and the segment does not intersect obstacles.
Hence, we can directly connect adjacent keypoint candidates by straight segments and a connectivity graph can be made by an iterative process of this operation.
Each node of a connectivity graph represents a keypoint candidate.
After we get a connectivity graph, we run a graph search algorithm such as Dijkstra or $\mathrm{A}^*$ to specify keypoints $k^*_1, \ldots, k^*_l$ (line 5).
If the distance between two adjacent of selected keypoints is less than $\varepsilon$, the keypoints are merged and we take the mid point of the two to reduce the number of keypoints (lines 17--22).  
At the time solving the graph search problem, we assign the weight of each edge as a length of a segment connecting adjacent nodes.

However, until this step, we only consider workspace information.
Generally, robots have their own configuration space and for actual motion planning problems, considering only workspace information is insufficient.
Then, we solve inverse kinematics problem to get corresponding configurations by using the keypoints from BSP.
After that, we run a sampling-based motion planner such as RRT or its variants to examine whether there is a feasible path by sequentially choosing selected keypoints (line 6).
In general, there are multiple solutions of an inverse kinematics problem, hence we run motion planner using all solutions of inverse kinematics solutions and examine the feasibility and path.
Since each subproblem is simple, solving it takes a few seconds.
If there is no feasible path, we remove the terminal node in the selected keypoints from a connectivity graph (lines 7-8).
\begin{algorithm}[t]
\small 
\algsetup{linenosize=\small}
\caption{Online Execution}
\begin{algorithmic}[1] 
\label{alg_online_execution}
\REQUIRE initial state $x_{\mathrm{init}}$, target state $x_{\mathrm{target}}$, environment $\mathcal{W}$
\WHILE{isTermination == False}
    \STATE $k_{i+1}^* \gets \mathrm{KeyPointEstimation}(k_{i}^*, x_{\mathrm{target}}, \mathcal{W})$
    \IF{$\| k_{i+1}^* - x_{\mathrm{target}}\|_2 < \delta$ }
        \STATE isTermination $\gets \mathrm{True}$
    \ENDIF
    \STATE keypoints $\gets \mathrm{Append}(k_{i+1}^*)$
\ENDWHILE
\FORALL{$k^*_i \in$ keypoints}
    \STATE $p_i(x|\mathcal{W}) \gets \mathrm{RunCVAE}(k^*_i, k^*_{i+1}, \mathcal{W})$
\ENDFOR
\STATE $p(x|\mathcal{W}) \gets \mathrm{Synthesize}(p_1(x|\mathcal{W}), \ldots, p_N(x|\mathcal{W}))$
\RETURN $p(x|\mathcal{W})$
\end{algorithmic}
\end{algorithm}
Otherwise, we examine the solution path (line 10).
A function ExaminePath examines 1) whether the path passes subspaces that the robot previously passed and 2) if multiple paths do not satisfy the criterion of 1), which path is shortest.
We use a connectivity graph to examine 1) and if the path satisfies 1), the edge connected to the node (keypoint) removes from the graph (line 12).
Besides, by following the criterion of 2), we reject other paths that do not satisfy the criterion of 1) (line 12).
Thus, we specify the keypoints and paths that connect two adjacent keypoints even robots have high degree of freedom.
Finally, we use them to train a neural network to estimate the keypoints and a CVAE to generate appropriately biased non-uniform sampling distribution.

\subsection{Online Execution (Algorithm \ref{alg_online_execution})}
After training (Algorithm \ref{alg_offline_learning}), we execute our framework in a given motion planning problem.
We firstly estimate the keypoints by the trained neural network using a given workspace information (lines 2--4).
At the time of estimation, we sequentially estimate the keypoint by using the keypoint at the current step $k_{i}^*$ (line 2).
We then make an array that contains keypoints (line 3).
We repeat this estimation process until the automatic termination criterion is satisfied (line 4).
In line 4, we raise the following criterion.
\begin{align}
    \| k_{i+1}^* - x_{\mathrm{target}} \|_2 < \delta
\end{align}
If the distance between the estimated keypoint and the target state $x_{\mathrm{target}}$ is less than appropriate chosen $\delta$, we cease the sequential estimation and move to CVAE execution (lines 8--10).

In line 9, we pass the two adjacent keypoints and environment information to CVAE and CVAE estimates non-uniform sampling distribution whose extremums are along the feasible path connecting the two keypoints.

Finally, we synthesize the local samplers provided by CVAE (line 11).
Unlike \cite{chamzas2019using, chamzas2021learning, chamzas2022learning}, we utilize the information of entire environment $\mathcal{W}$ to estimate non-uniform sampling distributions even when we consider subproblems.
This allows us to avoid the need to explicitly handle decomposition errors, thereby mitigating their potential impact on the sampling process.
We argue the details about decomposition error.
To begin with, we discuss the following approximation
\begin{align}
    p_i(x|\mathcal{W}) \approx p_i(x|\mathcal{L}_i).
\end{align}
where $\mathcal{L}_i$ is \textit{i}-th subspace. 
This approximation implies that the sampling distribution for a subproblem relies solely on the subspace in which the subproblem is defined. 
However, it is important to note that this approximation may lead to large decomposition errors.
For instance, in the robotic arm planning, the entire robot body may not be in a subspace.
In that case, considering collisions in a specific subspace is insufficient because the robot may collide with obstacles outside the subspace.
Therefore, to estimate the valid samples accurately, we should take entire environment into account.
For the above reason, we do not use this approximation.

We synthesize the local samplers by a weighted mean manner.
Given a set \textit{N} of local samplers $p_{i}(x|\mathcal{W})$, they are combined in the following way
\begin{align}
    p(x|\mathcal{W}) = \sum^N_{i=1} w_i p_{i}(x|\mathcal{W})
\end{align}
where $w_i$ denotes a weight for a corresponding sampling distribution $p_i(x|\mathcal{W})$.
To retain the probabilistic completeness of the sampling procedure, we still employ uniform sampler with a chance $\lambda \in [0,1]$ i.e. if $\lambda = 0$, we only use non-uniform samplers and if $\lambda = 1$, we do not use any non-uniform samplers.

\section{EXPERIMENTS}
In this section, we evaluate the performance of our framework in various motion planning problems.
To evaluate the performance, we applied our framework to 2D free-flying robot and 6DoF fixed robotic manipulator.
Although we handle these two robots in this paper, we believe our framework also works for other mobile or flying robots.

To evaluate the performance of our framework, we compare the performance with uniform, CVAE (Shortest Path) \cite{ichter2018learning}, and GMM with workspace decomposition \cite{chamzas2021learning}.
CVAE (Shortest Path) approach learns the probability distribution from the feasible solutions (typically shortest paths) to infer the valid configurations based on workspace information and does not use any decomposition techniques.
We use the implementation of CVAE provided by the authors of \cite{ichter2018learning} for CVAE (Shortest Path) and our framework.
We increase the dimension of latent space when we use the implementation for higher dimensional systems.  
As evaluation metrics, we consider planing time, cost, and number of valid samples generated from experience-based non-uniform samplers. 
We implement our framework in Python and run all experiments on an AMD Ryzen 9 machine with 3.30 GHz cores with 32.0 GB of memory.

\subsection{2D Free-Flying Robot}
\begin{figure}[t]
 \centering
    \subfigure[Environment 1]{%
        \includegraphics[clip, width=0.99\columnwidth]{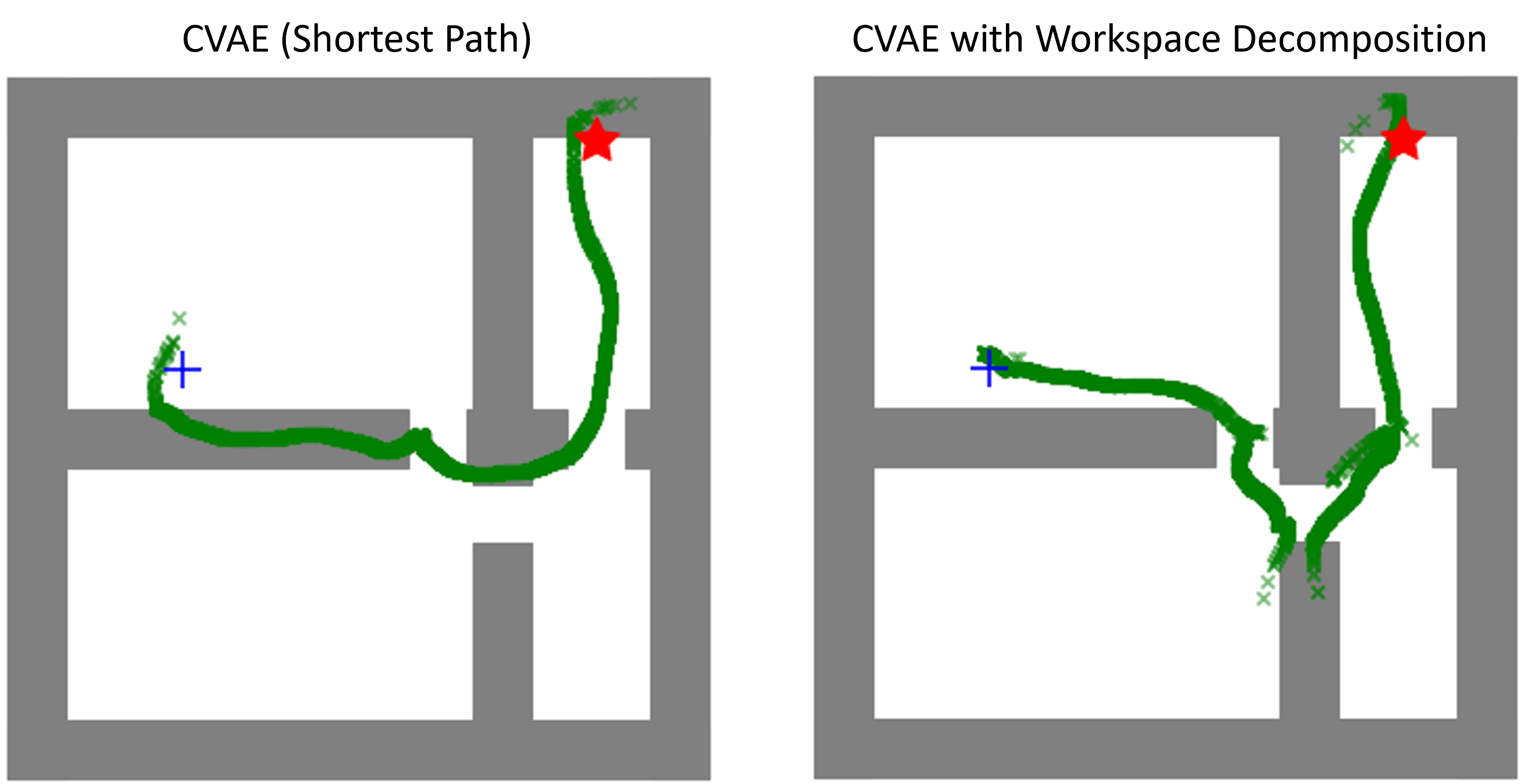}}
    \subfigure[Environment 2]{%
        \includegraphics[clip, width=0.99\columnwidth]{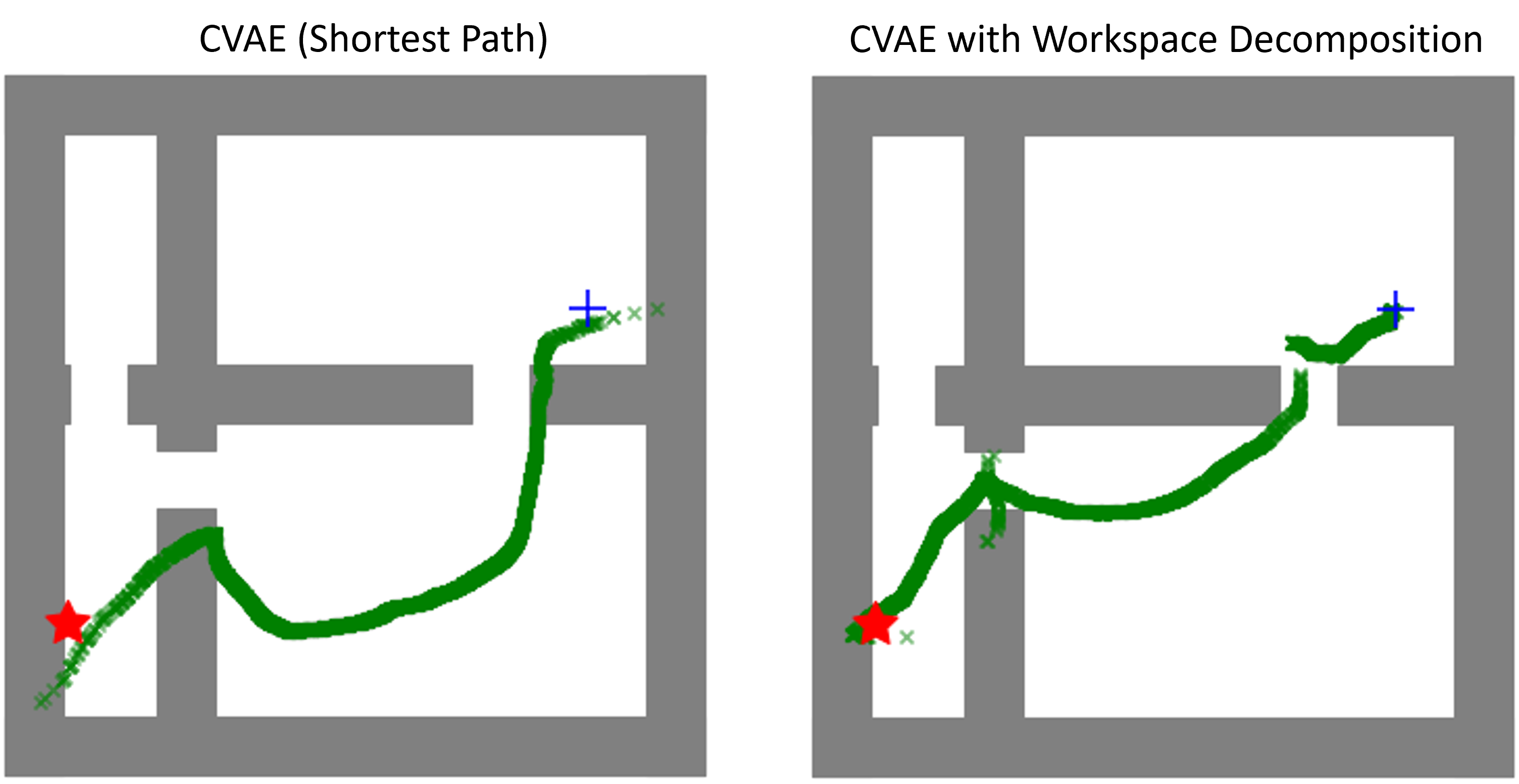}}
 \caption{A comparison of informed distributions. In the figures, the blue cross displays an initial state and the red star exhibits a target state. Green crosses are samples drawn from the informed distributions. The left figure displays an informed distribution without any workspace decomposition technique and the right figure exhibits an informed distribution with workspace decomposition. We can see that the number of valid samples increases by workspace decomposition.}
 \label{comparison_dist}
\end{figure}

To begin with, we handle problems with a 2D free-flying robot in a 2D maze.
We assume the free-flying robot as a point mass. 
The maze is constructed by several rectilinear walls and the gaps of these walls make some challenging regions (narrow passage) in the workspace (Fig. \ref{comparison_dist}).
When we apply our framework to the 2D maze, we can automatically extract the gaps of the rectilinear walls and acquire dense samples in the gaps.

To train a neural network for keypoint estimation, we use 12,000 randomized maze environments as training data.
For CVAE training, we use 8,000 randomized environments.
For a fair comparison, the training dataset for CVAE has solutions to the same randomized motion planning problems, and each contains the solutions to entire problems and the solutions to subproblems, respectively.
To analyze the performance of our framework, we compare the planning time, cost, and success rate with uniform sampling and CVAE (Shortest Path).

To begin with, we plot the samples drawn from a non-uniform sampling distribution inferred by CVAE (Shortest Path) and our approach in Fig. \ref{comparison_dist}.
In Fig. \ref{comparison_dist}, the blue cross is the initial state of the robot and the red star is the target state.
Green crosses are samples drawn from a non-uniform sampling distribution.
As shown in Fig. \ref{comparison_dist}, the non-uniform samples from CVAE (Shortest Path) are located in the walls and sometimes the distribution seems like simply connecting the initial and the target states. 
That means CVAE (Shortest Path) sometimes fails to take the conditions related to obstacles into account because of the high variance among the randomized queries.
On the other hand, most of the non-uniform samples from our approach are located in valid regions in the workspace.
This demonstrates the effectiveness of our approach that performs decoupling a problem into simple subproblems and make the variance among subproblems small. 

\begin{figure}[t]
 \centering
        \includegraphics[clip, width=0.95\columnwidth]{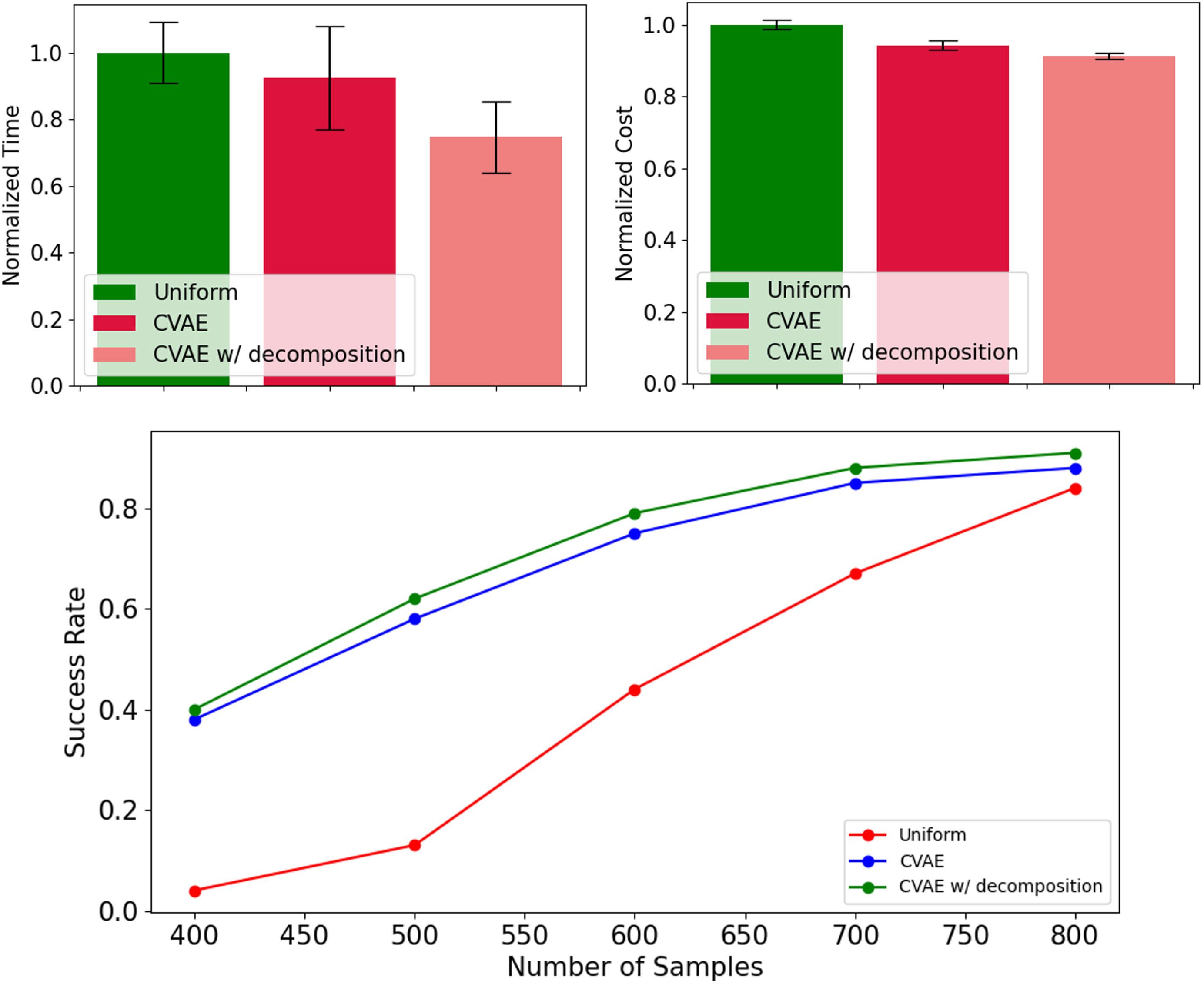}
 \caption{The analysis results of Environment 1. Each figure shows normalized planning time and planning cost with respect to number of non-uniform samples, respectively. The error bar indicates a 95\% confidence interval for a Gaussian distribution. As shown in these figures, workspace decomposition decreases the planning time and cost compared with CVAE case. Note that we normalize the planning time and cost with respect to those of uniform sampling.}
 \label{resultv-a-1}
\end{figure}
\begin{figure}[t]
 \centering
        \includegraphics[clip, width=0.95\columnwidth]{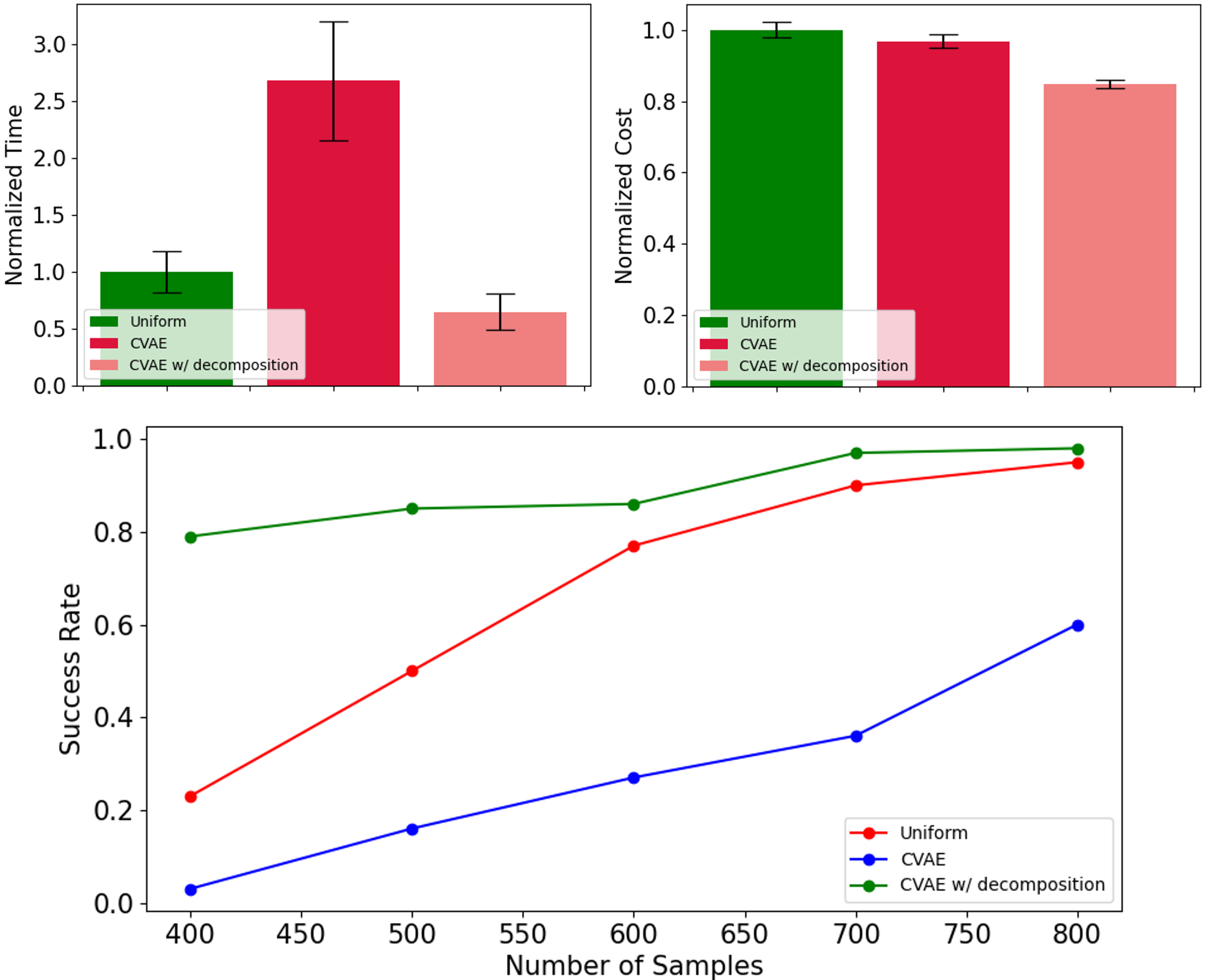}
 \caption{The analysis results of Environment 2. Each figure shows normalized planning time and planning cost with respect to number of non-uniform samples, respectively. The error bar indicates a 95\% confidence interval for a Gaussian distribution. As shown in these figures, workspace decomposition decreases the planning time and cost compared with CVAE case. Note that we normalize the planning time and cost with respect to those of uniform sampling.}
 \label{resultv-a-2}
\end{figure}
We then analyze the performance of our planner in the environment 1 and 2 in Fig. \ref{comparison_dist}.
We run RRT* to analyze the performance and utilize the 3000 non-uniform samples generated by CVAE.
For evaluation, we run the simulation 100 times in each environment and take the mean of planning time and cost as performance indices.
We also analyze success rate when sampling 400, 500, 600, 700, and 800 samples.
Fig. \ref{resultv-a-1} and Fig. \ref{resultv-a-2} display the results of the analyses. 
Both planning time and cost, our planner demonstrates better performance compared with other baselines.
For success rate, our planner achieves higher success rate even if the number of samples is small.

\subsection{6D Robotic Manipulator}
In this section, we address 6D robotic manipulator planning scenarios, which are more complex than 2D cases.
We consider a real shelf environment.
In this problem, the robotic arm inserts its body into the shelf with obstacles, and the initial states, the target states, the position of the shelf, and the positions of the obstacles inside the shelf are randomized.
For the real shelf environment, the shelf has three levels and each level's vertical width is small.
Hence, each level creates narrow regions and the presence of obstacles on the shelf also makes this problem more challenging.

We use 1,000 randomized environment as training dataset for the neural network of keypoint estimation.
Besides we utilize 500 randomized environments as training dataset for CVAE training and the database for GMM approach.
Same as the 2D maze case, the dataset for CVAE (Shortest Path), GMM, and our approach is the same. 
To validate the performance of our motion planner, we use RRT* and we use 7,000 non-uniform samples generated by CVAE.
\begin{figure}[t]
 \centering
    \includegraphics[clip, width=0.99\columnwidth]{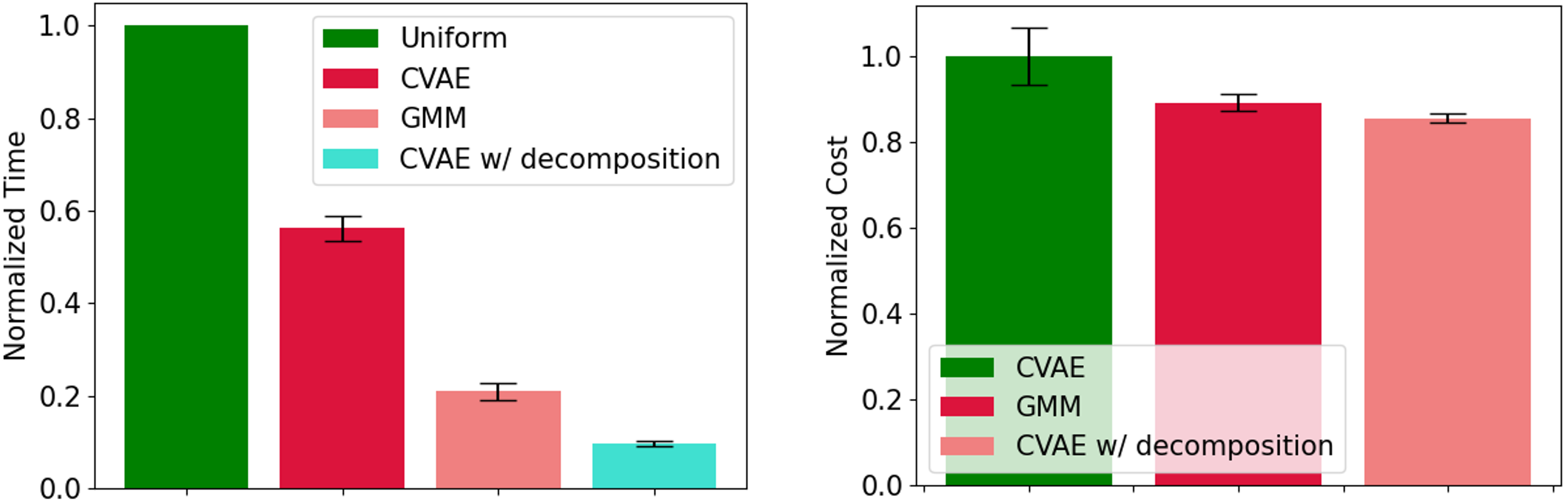}
 \caption{The analysis results of real shelf environment. Each figure shows normalized planning time and planning cost, respectively. The error bar indicates a 95\% confidence interval for a Gaussian distribution. As shown in these figures, workspace decomposition decreases the planning time and cost compared with CVAE and GMM case. Compared to 2D example, the reduction rate becomes larger. Note that we normalize the planning time with respect to those of uniform sampling. Since uniform sampling reaches timeout, we normalize the cost with respect to that of CVAE.}
 \label{resultv-b}
\end{figure}

We plot the analysis results of planning time and planning cost versus number of samples in Fig. \ref{resultv-b}.
For evaluation, we run the simulation 10 times and take the mean as performance indices.
We use 7,000 samples from non-uniform sampling distributions.
For the weights of local samplers for the GMM planner, we assign equal weights as presented in \cite{chamzas2019using, chamzas2021learning, chamzas2022learning}.
Both planning time and cost, our planner shows the better performance even for high variance in environments.
Besides, we use relatively small number of dataset to achieve the performance.
These results demonstrate our framework is effective even in complex and realistic scenarios.

\section{CONCLUSION}
In this study, we presented a novel framework that combines workspace decomposition and experience-based non-uniform sampling to achieve an efficient motion planning process. 
Our proposed framework incorporates two neural network structures.
Firstly, a neural network is employed to estimate keypoints, which serve the purpose of decoupling motion planning problems and extracting challenging regions within the workspace by learning from BSP results.
Secondly, CVAE is utilized to estimate suitable non-uniform sampling distributions for each subproblem, utilizing the estimated keypoints.
Simulation studies conducted on a free-flying robot and a 6-DoF robotic arm demonstrate the effectiveness of our framework. 
The unification of workspace decomposition via BSP and non-uniform sampling is shown to accelerate the planning process and reduce costs in comparison to existing planners.

\bibliographystyle{IEEEtran}
\bibliography{root.bbl}	
\vspace{12pt}

\end{document}